# Efficient Triple Modular Redundancy for Reliability Enhancement of DNNs Using Explainable AI

Kimia Soroush
School of Electrical and Computer Engineering
Shiraz University.
Shiraz, Iran
kimiasoroush@hafez.shirazu.ac.ir

Nastaran Shirazi
School of Electrical and Computer Engineering
Shiraz University.
Shiraz, Iran
nastaran@hafez.shirazu.ac.ir

Mohsen Raji
School of Electrical and Computer Engineering
Shiraz University.
Shiraz, Iran
mraji@shirazu.ac.ir

*Abstract*— Deep Neural Networks (DNNs) are widely employed in safety-critical domains, where ensuring their reliability is essential. Triple Modular Redundancy (TMR) is an effective technique to enhance the reliability of DNNs in the presence of bit-flip faults. In order to handle the significant overhead of TMR, it is applied selectively on the parameters and components with the highest contribution at the model output. Hence, the accuracy of the selection criterion plays the key role on the efficiency of TMR. This paper presents an efficient TMR approach to enhance the reliability of DNNs against bit-flip faults using an Explainable Artificial Intelligence (XAI) method. Since XAI can provide valuable insights about the importance of individual neurons and weights in the performance of the network, they can be applied as the selection metric in TMR techniques. The proposed method utilizes a low-cost, gradient-based XAI technique known as Layer-wise Relevance Propagation (LRP) to calculate importance scores for DNN parameters. These scores are then used to enhance the reliability of the model, with the most critical weights being protected by TMR. The proposed approach is evaluated on two DNN models, VGG16 and AlexNet, using datasets such as MNIST and CIFAR-10. The results demonstrate that the method can protect the AlexNet model at a bit error rate of $10^{-4}$, achieving over 60% reliability improvement while maintaining the same overhead as state-of-the-art methods.

## I. INTRODUCTION

In recent years, Deep neural networks (DNNs) play a crucial role in various electronic devices and services, including smartphones, home appliances, drones, robots, and autonomous vehicles. Ensuring the reliability of these systems is critical to prevent malfunctions or failures. A commonly used approach to enhance hardware fault tolerance is Triple Modular Redundancy (TMR), which involves triplicating modules and determining the output through majority voting [1]. However, this method significantly increases hardware overhead, consuming 200% more resources, making it impractical for power-intensive machine learning accelerators. Moreover, not all components of a neural network need to be triplicated. Granularity plays a vital role, as certain CNN filter weights are more critical than others [2]. Therefore, it is essential to develop methods for evaluating the importance of various components within a neural network at different levels of granularity. Since different parts of a network contribute to accuracy in distinct ways, understanding its structure is key to designing more effective optimization strategies. In this context, identifying the most critical elements of the network becomes a priority.

Explainable Artificial Intelligence (XAI), an emerging field dedicated to interpreting AI systems, offers valuable tools for this purpose. By using XAI, it is possible to analyze CNNs and determine which neurons, weights, and features play the most significant roles in the overall performance of the network [3].

Prior research has explored various methods to enhance the reliability of neural network inference. Sabih et al. [7] utilized XAI to assign TMR to a subset of feature map elements based on their importance, scaled by layer sensitivities. Their approach uses DeepLIFT, a data-dependent XAI method. However, [6] takes a different approach by ranking neurons based on average output magnitudes and applying TMR uniformly to a fixed percentage (pNeu) of neurons, without considering the varying levels of criticality within the subset. This method may fail to account for the dynamic and context-specific importance of neurons under different input distributions or adversarial conditions. Boreta et al. [5] focuses primarily on evaluating fault tolerance concerning stuck-at faults in activation channels and introduces a TMR technique at the neuron level. However, it does not address a broader range of fault types, such as single-bit flips in weights, which can impact performance differently. Furthermore, its lack of granularity results in a 56% overhead. In [4], robustness against Single Event Upset (SEU)-induced parameter perturbations is measured. While this approach identifies sensitive bits (e.g., exponent bits) and protects them, it leaves other bits in the same parameter vulnerable, and its exhaustive bit-flipping search causes significant computational overhead. Thus, a method that protects all bits of critical parameters is needed. In [8], uniform fault injection is employed to randomly introduce faults across all neural network layers, identifying parameters highly susceptible to soft errors. However, this method introduces considerable overhead and focuses on parameter-level vulnerability without deeply analyzing or leveraging sensitivity information for specific layers or features. Additionally, the framework is tailored for binary neural networks (BNNs), limiting its applicability to other neural network types that rely on higher-precision weights and activations.

In general, prior works have explored the use of TMR techniques to protect neural networks from various types of bit flips. However, these methods often result in significant overhead due to insufficient granularity. This lack of granularity leads to unnecessary redundancy for less critical components while failing to provide enough protection for the most critical ones. Additionally, many approaches rely on data-dependent metrics to identify the most critical weights. As a result, these methods are inefficient and highly sensitive to changes in input data, which makes their reliability techniques ineffective in dynamic scenarios. Furthermore, the metrics used in these approaches are often computationally exhaustive. Most methods identify critical components by performing exhaustive fault injection or bit-flipping across all weights, which is resource-intensive and impractical for large-scale networks. A more efficient and data-agnostic approach is needed to accurately and effectively identify and protect the most important components without excessive overhead.

In this paper, we propose using TMR as a technique to protect neural network parameters against bit-flip faults. To minimize the overhead of TMR, it is crucial to adopt an approach that identifies the most critical parameters efficiently, avoiding exhaustive computations and reliance on data-dependent metrics. Our contributions are as follows:

- Use an XAI method, Layer-wise Relevance Propagation (LRP), to calculate the importance scores of network weights.
- Apply TMR selectively to enhance the DNN reliability by protecting the most critical weights against bit flip faults.

By combining these techniques, our approach effectively identifies and protects the critical parameters with minimal overhead. The proposed approach is evaluated on two DNN models, VGG16 and AlexNet, using datasets such as MNIST and CIFAR-10. The results demonstrate that the method can protect the AlexNet model at a bit error rate of $10^{-4}$, achieving over 60% reliability improvement while maintaining the same overhead as state-of-the-art methods.

The remainder of this paper is organized as follows: Section 2 provides some background and section 3 presents the proposed framework, detailing the integration of LRP and TMR. Section 4 describes the experimental setup and results, while Section 5 concludes the paper and outlines future research directions.

## II. BACKGROUND

This section provides background on LRP, fault injection, and TMR to help understand the rest of the paper.

### A. Layer-wise Relevance Propagation (LRP)

LRP is an explainability method used to attribute the output of the model to its inputs. In the context of DNNs, LRP assigns relevance scores to individual neurons, weights, or input features, reflecting their contribution to the output. These scores can guide the identification of critical components within the network.

Determining the relevance of individual weights begins with assessing the relevances of neurons. To delve deeper into the process of calculating neuron importance scores, we focus on a specific output neuron, denoted as $j$. The importance score assigned to the neuron $j$, denoted as $R_j$, is distributed among its lower-layer input neurons $i$ using the formula defined in the following formula:

$$R_{i \leftarrow j} = \frac{z_{ij}}{z_j} R_j \qquad \text{Eq. (1)}$$

where $z_{ij}$ represents the contribution of neuron $i$ to the activation of neuron $j$, $z_j$ is the sum of pre-activations $z_{ij}$ at output neuron $j$, i.e.:

$$z_j = \sum_j z_{ij} \qquad \text{Eq. (2)}$$

To ensure the conservation principle is upheld for all $i$ contributing to $j$, we have the following relation:

$$\sum_i R_{i \leftarrow j} = R_j \qquad \text{Eq. (3)}$$

where the importance score is represented as $R_{i \leftarrow j}$, it indicates the extent to which the effect of output neuron $j$, assigned importance score $R_j$, is propagated to its lower layer input neurons $i$, following the decomposition rule mentioned above.

By using the direct connection to network performance LRP and its straightforward integration with different network layers, our reliability enhancement method becomes both interpretable and scalable.

### B. Fault Injection

Fault injection is a technique to emulate hardware faults, such as flipping a bit in the binary representation of a weight. For instance, flipping bit 29 in a 32-bit floating-point number introduces significant perturbations to the weight value as shown in Figure 1. This figure demonstrates the effect of a bit-flip error on a neural network weight. A single bit-flip transforms the original binary value (0101) to a corrupted value (1101), highlighting how hardware faults can significantly alter weight representations and potentially disrupt the network's performance and reliability.

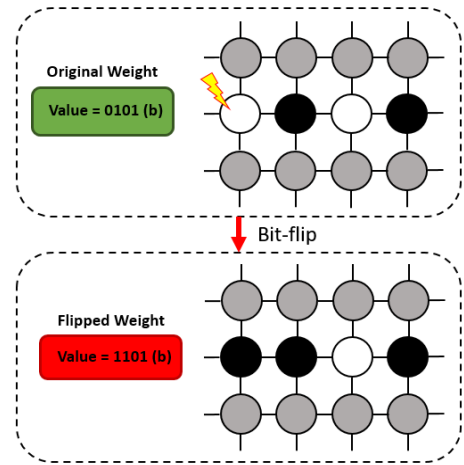

Figure 1: Illustration of a bit-flip fault in a neural network weight matrix: the original weight (value = 0101 in binary) is altered to a flipped weight (value = 1101 in binary).

### C. Triple Modular Redundancy (TMR)

TMR enhances the reliability of a given DNN by triplicating critical model parameters and employing majority voting to recover from single faults. It is particularly effective for protecting critical parameter of a DNN identified through an evaluation technique like LRP. Figure 2 illustrates an error mitigation mechanism for neural network weights. The original weight is replicated, and all three versions are fed into a voter module. If a fault, such as a bit-flip, occurs in one of the parameters (original or replica), the voter determines the correct weight using a majority voting strategy. This approach ensures resilience against single-bit faults, improving reliability in DNN models.

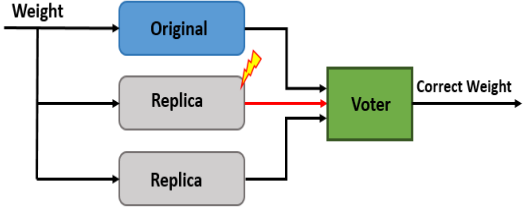

Figure 2: Error mitigation using weight replication and majority voting for reliability enhancement.

### III. PROPOSED METHOD

In this section, we present the proposed efficient TMR approach to enhance the reliability of DNNs against bit-flip faults. To minimize the overhead commonly associated with TMR, we utilize XAI techniques, specifically LRP, to identify the most critical weights within the network. By selectively applying TMR to these critical weights, we achieve a balance between reliability and computational efficiency. In the following, we describe the steps of the proposed method, which is also outlined in Algorithm 1 through 4 steps.

**Step1: Identify critical weights using LRP:** To avoid applying TMR blindly across all weights, we use LRP to compute the importance scores of individual weights. For each layer of the DNN, the relevance score of each weight is calculated using a predefined LRP rule. These scores quantify the contribution of each weight to the overall network output. Once the scores are obtained, the weights are sorted in descending order based on their importance. The top 1% of the weights with the highest relevance scores are identified as critical components, forming the subset $W_{top}$. This selective identification allows us to focus on protecting the most sensitive parts of the network, significantly reducing the redundancy overhead associated with TMR (lines 1 to 7 of Algorithm1).

LRP works by propagating the output relevance RRR backward through the network. The relevance at each neuron is redistributed among its inputs based on a set of rules [10,11]. The most commonly used LRP rules include:

- **LRP-0**: For a neuron i, its relevance $R_i$ is distributed to its inputs j:

$$\mathcal{R}_j = \sum_i \frac{a_j w_{ji}}{\sum_k a_k w_{ki}} \mathcal{R}_i \qquad \text{Eq. (4)}$$

The 0-rule is applied to ReLU layers because it focuses only on positive contributions, making the results clearer.

- **LRP-$\epsilon$**: To prevent numerical instability when small activations dominate, a stabilizer $\epsilon$ is added:

$$\mathcal{R}_j = \sum_i \frac{a_j w_{ji}}{\sum_k a_k w_{ki} + \epsilon} \mathcal{R}_i \qquad \text{Eq. (5)}$$

The ε-rule is applied to fully connected layers as it avoids errors by preventing division by very small numbers.

- **LRP-$\gamma$**: This rule is designed to provide a relevance score for each component of a neural network, emphasizing positive contributions of weights while suppressing negative contributions:

$$\mathcal{R}_j = \sum_i \frac{a_j(w_{ji} + \gamma \cdot \max(w_{ji},0))}{\sum_k a_k(w_{ki} + \gamma \cdot \max(w_{ki},0)) + \epsilon} \mathcal{R}_i \qquad \text{Eq. (6)}$$

The γ-rule is applied to convolutional layers since it increases positive relevance to highlight important features.

By applying these rules to different layers, LRP can identify the most critical weights in the network, specifically those with the highest relevance scores. In this research, each formula is customized to each layer based on its specific characteristics.

**Step2: Apply TMR to critical weights:** For the weights identified in $W_{top}$, TMR is applied to protect them from bit-flip errors. Each critical weight is replicated into three copies, referred to as $w_k^{(1)}$, $w_k^{(2)}$, and $w_k^{(3)}$. During inference, the effective weight is computed using a majority voting mechanism. This ensures that even if a single replica experiences a bit-flip error, the effective weight remains robust to the fault. After sorting the weight relevances calculated in the previous step, limiting TMR to only 1% of the most critical weights significantly reduces the computational and storage overhead compared to applying TMR to all weights in the network (lines 8 to 11 of Algorithm1).

Mathematically, the relevance of weights is calculated using LRP-based formulas, as described in the background. The probability of a fault occurring at a random bit (e.g., rnd) is represented as:

$$P(f_{rnd}) = \frac{NBF}{NT} \qquad \text{Eq. (7)}$$

where NBF=number of bits flipped, NT=number of total weights. For weights $w_1, w_2, w_3$ (TMR replicas), the effective weight is given by:

$$w_{eff} = MajorityVote(w_1, w_2, w_3) \qquad \text{Eq. (8)}$$

By combining LRP for critical weight identification and TMR for protection, this methodology offers a robust framework for analyzing and improving DNN resilience in fault-prone environments.

**Step3: Inject faults:** To evaluate the robustness of the enhanced model, fault injection is simulated by introducing bit-flips to the critical weights based on a specified Bit Error Rate (BER). A bit-flip is simulated by flipping a specific bit in the binary representation of the weight. This step allows us to assess the fault tolerance of the model under real-world error conditions (lines 12 to 14 of Algorithm1).

**Step4: Evaluate performance:** Finally, the accuracy of the model under fault conditions is evaluated. The proposed approach demonstrates its ability to maintain high accuracy levels even in the presence of bit-flip errors, validating the effectiveness of selective TMR in conjunction with LRP (line 15 of Algorithm1).

**Algorithm 1** Efficient TMR with LRP for CNN Reliability Enhancement
**Input:** Model $M$, LRP Rules $R$, Bit Error Rate (BER)
**Output:** Enhanced Model with TMR and Accuracy under Fault Injection
1: **for** each layer $L$ in $M$ **do**
2:    **for** each weight $w_{ij}$ in $L$ **do**
3:       Compute relevance score: $S_{ij}$ = LRP Rule($w_{ij}$)
4:    **end for**
5: **end for**
6: Sort weights by relevance scores: $W_{sorted}$ = sort($W, S,$ desc)
7: Select top 1% most critical weights: $W_{top} = W_{sorted}[:\lceil 0.01 \times |W| \rceil]$
8: **for** each weight $w_k \in W_{top}$ **do**
9:    Create three replicas: $w_k^{(1)}, w_k^{(2)}, w_k^{(3)}$
10:   Compute effective weight with majority voting:
$$w_{eff} = \text{MajorityVote}(w_k^{(1)}, w_k^{(2)}, w_k^{(3)})$$
11: **end for**
12: **for** each weight $w_k \in W_{top}$ **do**
13:   Introduce bit-flip based on BER: $w_k^{flipped} = w_k \oplus (1 \ll 29)$
14: **end for**
15: Accuracy$_{faulted}$ = evaluate($M$)
16: **Return** Enhanced Model with TMR and Accuracy$_{faulted}$

## IV. EXPERIMENTAL RESULTS

To demonstrate the efficacy of the proposed reliability enhancement approach, we conducted a series of experiments to investigate the role of XAI in identifying critical weights for applying the TMR technique. The following section details the experimental setup, followed by a comprehensive analysis to identify the most critical weights using different methods. Also, the hardware overhead of each method is calculated and presented to evaluate the efficiency of the proposed method from a hardware perspective.

### A. Experiment Setup

The experiments are conducted using two DNNs, VGG-16 and AlexNet, both pre-trained on the ImageNet dataset. The proposed framework is implemented with PyTorch and torchvision libraries. The experiments are executed on a system featuring an Intel(R) Core i9-10900K 3.70GHz processor and an NVIDIA GeForce RTX GPU with 12GB of memory, ensuring efficient GPU utilization. The performance of the reliability enhancement techniques is evaluated on two publicly available datasets, CIFAR10 and MNIST.

### B. Sensitivity Analysis

In order to investigate the efficacy of XAI in identifying the most critical weights, we employed different metrics in identifying the weights for fault injection; i.e. injecting faults on the most critical weight leads to more accuracy degradation and higher loss in the model.

Figure 3(a) compares the accuracy degradation of AlexNet on MNIST dataset under bit-flip fault injection (FI) using three methods: 1) Random FI [4] in which the weights are randomly selected for injecting faults, 2) Magnitude-based Fault Injection (FI) [5], where faults are injected into weights with higher magnitudes, is based on the intuition that weights with greater magnitudes have a higher impact on the output of the model, and 3) XAI-based FI in which the importance scores obtained through LRP is used to find the weights to inject faults.

The results show that XAI-based FI causes the most significant accuracy drop, as it targets the most critical weights identified by their contribution to the output of the network. In contrast, magnitude-based FI leads to moderate degradation, while random FI has the least impact, with accuracy decreasing gradually. Overall, the figure highlights the sensitivity of critical weights to bit flips. In Figure 3(b), the loss value is analyzed under the same bit error rate injections as in Figure 3(a) XAI-based, magnitude-based, and random FI. The results demonstrate a similar trend, where the XAI-based FI method results in the sharpest increase in loss values as the bit error rate rises, reflecting the significant impact of flipping the most critical weights on the performance of the network. The magnitude-based method causes a moderate increase in the loss, indicating that bit flips injected into high-magnitude weights still damages the network performance but less than XAI-targeted weights. Meanwhile, the random FI method shows the smallest and most gradual increase in loss, as the faults are distributed randomly, reducing their overall impact. Together, Figures 3(a) and 3(b) emphasize the importance of identifying and protecting critical weights to minimize both accuracy degradation and loss increase in the presence of bit flips.

Figure 4(a) compares the accuracy degradation of the VGG16 on the CIFAR-10 dataset under bit flip fault injection using three methods. The results show that XAI-based bit flips cause the most significant accuracy drop, as they target the most critical weights determined by their contribution to the network's output. In contrast, magnitude-based FI lead to moderate degradation, as they focus on high-magnitude weights without considering their overall importance. Random FI have the least impact, with accuracy decreasing gradually due to the distributed and non-specific nature of the injected faults. Overall, the figure highlights the heightened sensitivity of critical weights in VGG16 to bit flips, particularly those identified by XAI.

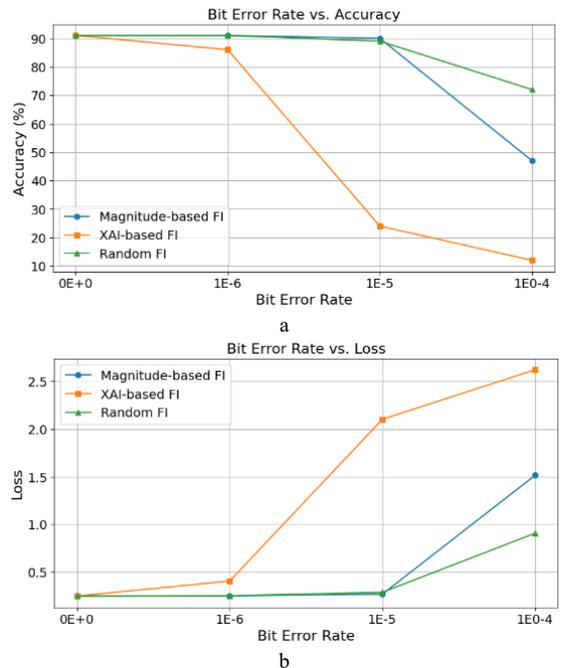

Figure 3 a) Comparison of accuracy degradation versus bit error rate for different fault injection methods, b) Comparison of loss value versus bit error rate for different fault injection methods on AlexNet.

In Figure 4(b), the loss value is analyzed for the same FI methods applied to the VGG16 network on the CIFAR-10 dataset. The results demonstrate a similar pattern, with the XAI-based method leading to the sharpest increase in loss values as the bit error rate rises, underscoring the severe impact of targeting highly influential weights. The magnitude-based FI method results in a moderate increase in loss, reflecting the disruption caused by flipping high-magnitude weights, which are less critical than those targeted by LRP. Random FI exhibit the smallest and most gradual increase in loss, as the faults are uniformly distributed without targeting specific weights. Together, Figures 4(a) and 4(b) highlight the critical need for robust fault tolerance mechanisms to protect important weights and maintain the reliability and performance of VGG16 in the presence of bit flips.

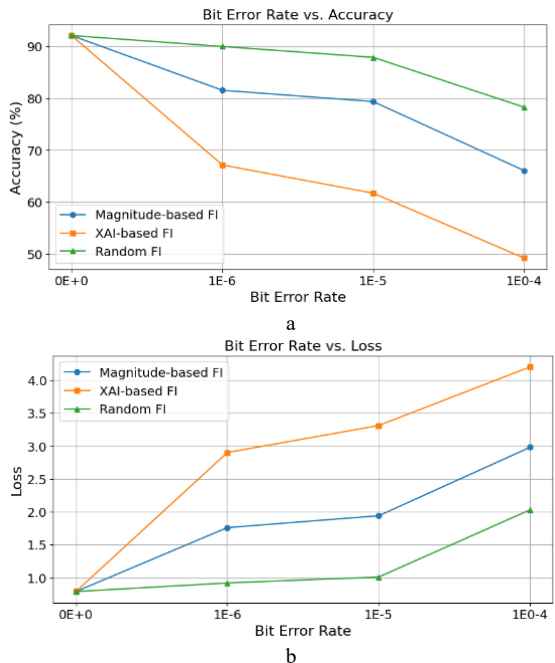

Figure 4: a) Comparison of accuracy degradation versus bit error rate for different fault injection methods, b) Comparison of loss value versus bit error rate for different fault injection methods on VGG16.

### C. Reliability Enhancement using TMR

In order to study the efficacy of the proposed XAI-based TMR in terms of reliability enhancement, we study the reliability enhancement achieved with different TMR approaches on DNN models. To this end, TMR technique is applied to 1% of the most critical weights identified with different approaches, including randomly selection [4], selecting based on weight magnitude [5], selecting based on LRP relevance scores. The obtained results on AlexNet on the MNIST dataset is shown in Figure 5.

As shown in Figure 5(a), after injecting bit flips at various error rates, the accuracy of the modified model is measured. The XAI-based TMR method maintains high accuracy, with at most a 2% accuracy degradation, demonstrating its effectiveness in identifying the most important weights. In contrast, the magnitude-based TMR performs significantly worse, with accuracy dropping to less than 30%, as it ranks weights based just on their values, which is not a reliable metric for identifying critical weights. In random TMR, 1% of the weights are randomly selected and tripled, leading to even lower performance. Therefore, we can conclude that the XAI-based TMR effectively identifies and protects the most important weights, outperforming the other methods. Figure 5(b) presents the same experiment conducted on VGG16 with the CIFAR-10 dataset. The results further confirm that the XAI-based TMR effectively selects the most important weights for TMR, consistently outperforming alternative methods.

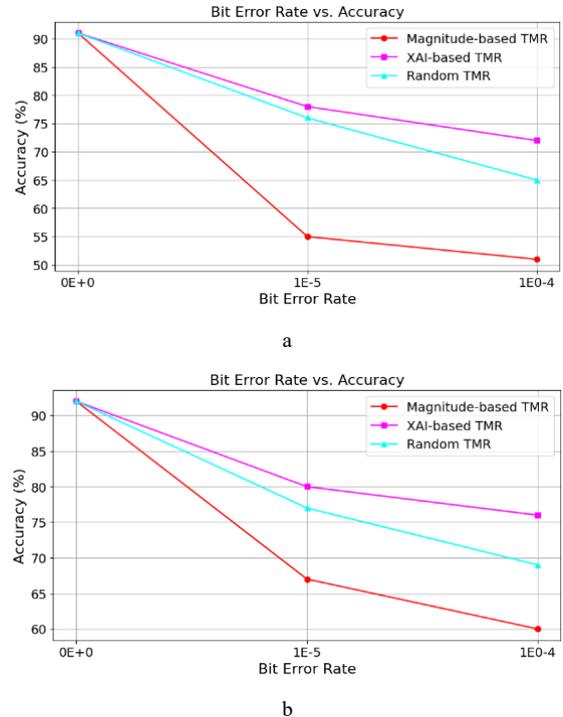

Figure 5: a) Comparison of accuracy degradation versus Bit Error Rate under the impact of Triple Modular Redundancy (TMR) on the 1% most important weights for different methods on AlexNet. b) Comparison of accuracy degradation versus Bit Error Rate under the impact of Triple Modular Redundancy (TMR) on the 1% most important weights for different methods on VGG16.

### D. Overhead Analyis of Different TMR approaches

In order to show the efficacy of proposed XAI-based TMR in terms of memory overhead, we study the models size obtained after applying TMR based on different methods on DNN models.

Figures 6(a) and 6(b) present the results, where the x-axis represents the methods applied, and the y-axis shows the corresponding model size. Both experiments were conducted at a bit error rate (BER) of 1E-4, targeting an accuracy loss of 5% on AlexNet model with MNIST dataset, and VGG16 model with CIFAR10 dataset respectively. The model size of AlexNet is 242MB and VGG16 model size is 522MB. We evaluated three methods for implementing TMR: Full TMR in which all weights are tripled [12], Magnitude-based TMR in which the weights with higher magnitude are chosen to be tripled [5], and XAI-based TMR in which the LRP-guided method is used to choose the weights to be tripled. In these experiments, random fault injections were performed at the specified BER, progressively increasing until the accuracy of the network degraded up to 5%. The full TMR method triples all weights in the network, significantly increasing the model size. In contrast, the weight magnitude-based and LRP-guided approaches selectively triple weights based on their importance rankings.

According to figure 6(a), it is evident that for full TMR approach, each weight requires an additional 8 bytes, tripling the memory allocation for the entire network, and it leads to a 200% overhead on model size. In comparison, the weight magnitude-based method selects weights for TMR based on their magnitudes and has an overhead of 21%. By avoiding redundant duplication of less critical weights, it causes a lower overhead than the full TMR method. The XAI-based approach demonstrates the least memory overhead among the three methods, with an overhead of only 1%. By focusing on the most influential weights, XAI-based method minimizes the additional memory requirement while still maintaining resilience against faults. These results highlight the advantages of selective TMR methods, particularly the XAI-based approach, in achieving fault tolerance with minimal memory overhead. According to Figure 6(b), which presents results for the VGG16 model on the CIFAR10 dataset, a similar trend is observed. The full TMR approach again causes a significant overhead and it results in a 200% increase in model size. In contrast, the weight magnitude-based method applies selective TMR based on weight magnitudes and has an overhead of 16%, reducing overhead by avoiding unnecessary duplication of less impactful weights. Notably, the XAI-based approach achieves the most efficient memory utilization, with only a 1% overhead. This highlights its ability to selectively identify and triple only the most critical weights, ensuring both robustness and minimal resource consumption. These findings reinforce the benefits of the XAI-based method as a superior selective TMR approach, effectively balancing fault tolerance and memory efficiency across different models and datasets.

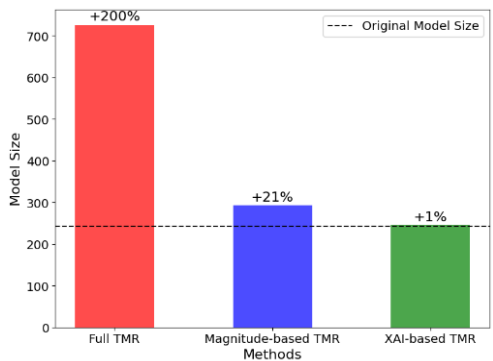
a

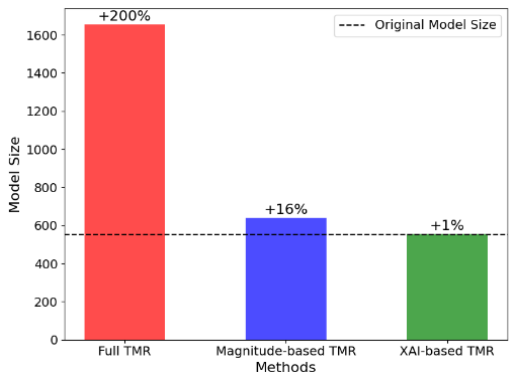
b

Figure 6: a) Comparison of overhead caused by TMR using three different methods on AlexNet. b) Comparison of overhead caused by TMR using three different methods on VGG16.

## V. CONLCUSION

In this paper, we presented a method to enhance the reliability of DNNs against bit-flip faults. We selectively applied TMR to protect the most critical components, reducing overhead while improving reliability. To handle the significant overhead of TMR, we utilized XAI to identify the most important weights in the network. Specifically, we employed LRP, a low-cost XAI technique, to calculate importance scores for the weights. We tested our method on VGG16 and AlexNet using the MNIST and CIFAR-10 datasets. The results showed that our approach improved the reliability of AlexNet by more than 60% at a bit error rate of $10^{-4}$ while maintaining the same overhead as other advanced methods. This work shows that XAI can be used to make fault-tolerance techniques more efficient. Future studies can explore this approach on other network types and under different conditions, such as other type of bit flip faults.